\documentclass[letterpaper]{article} 
\usepackage{aaai23}  
\usepackage{times}  
\usepackage{helvet}  
\usepackage{courier}  
\usepackage[hyphens]{url}  
\usepackage{graphicx} 
\usepackage{natbib}  
\usepackage{caption} 
\usepackage{algorithm}
\usepackage{algorithmic}
\usepackage{amsmath}
\usepackage{booktabs}
\usepackage{multirow}
\usepackage{xcolor}

\usepackage{amssymb}

\usepackage{newfloat}
\usepackage{listings}
\DeclareCaptionStyle{ruled}{labelfont=normalfont,labelsep=colon,strut=off} 
\floatstyle{ruled}
\newfloat{listing}{tb}{lst}{}
\floatname{listing}{Listing}
\title{TransVCL: Attention-enhanced Video Copy Localization Network \\ with Flexible Supervision}
\author{
    Sifeng He\textsuperscript{\rm 1}, Yue He\textsuperscript{\rm 1}, Minlong Lu\textsuperscript{\rm 1}, Chen Jiang\textsuperscript{\rm 1}, Xudong Yang\textsuperscript{\rm 1}, \\
    Feng Qian\textsuperscript{\rm 1*}, Xiaobo Zhang\textsuperscript{\rm 1}, Lei Yang\textsuperscript{\rm 1}, Jiandong Zhang\textsuperscript{\rm 2}
}
\affiliations{
    \textsuperscript{\rm 1}Ant Group     \textsuperscript{\rm 2} Copyright Protection Center of China \\

    \{sifeng.hsf, youzhi.qf\}@antgroup.com
}

\usepackage{bibentry}

\begin{document}

\maketitle

\begin{abstract}
Video copy localization aims to precisely localize all the copied segments within a pair of untrimmed videos in video retrieval applications. 
 Previous methods typically start from frame-to-frame similarity matrix {generated} by cosine similarity between frame-level features of {the} input video pair, and then detect and refine the boundaries of copied segments on similarity matrix under temporal constraints.
In this paper, we propose TransVCL: an attention-enhanced video copy localization network, which is optimized directly from initial frame-level features and trained end-to-end with three main components: a customized Transformer for feature enhancement, a correlation and softmax layer for similarity {matrix} generation, and a {temporal alignment module} for copied segments localization. 
In contrast to previous methods {demanding} the handcrafted similarity matrix, TransVCL incorporates long-range temporal information between feature {sequence pair} 
using self- and cross- attention layers. With the joint design and optimization of three components, the similarity matrix can be learned to present more discriminative copied patterns, leading to significant improvements over previous methods on segment-level labeled datasets (VCSL and VCDB). 
Besides the state-of-the-art performance in fully supervised setting, the attention architecture facilitates TransVCL to further exploit unlabeled or simply video-level labeled data.
Additional experiments of supplementing video-level labeled datasets including SVD and FIVR reveal the high flexibility of TransVCL from full supervision to semi-supervision (with or without video-level annotation).
Code is publicly available at https://github.com/transvcl/TransVCL.
\end{abstract}

\section{Introduction}

\begin{figure}[t]
\centering
\includegraphics[width=0.95\linewidth]{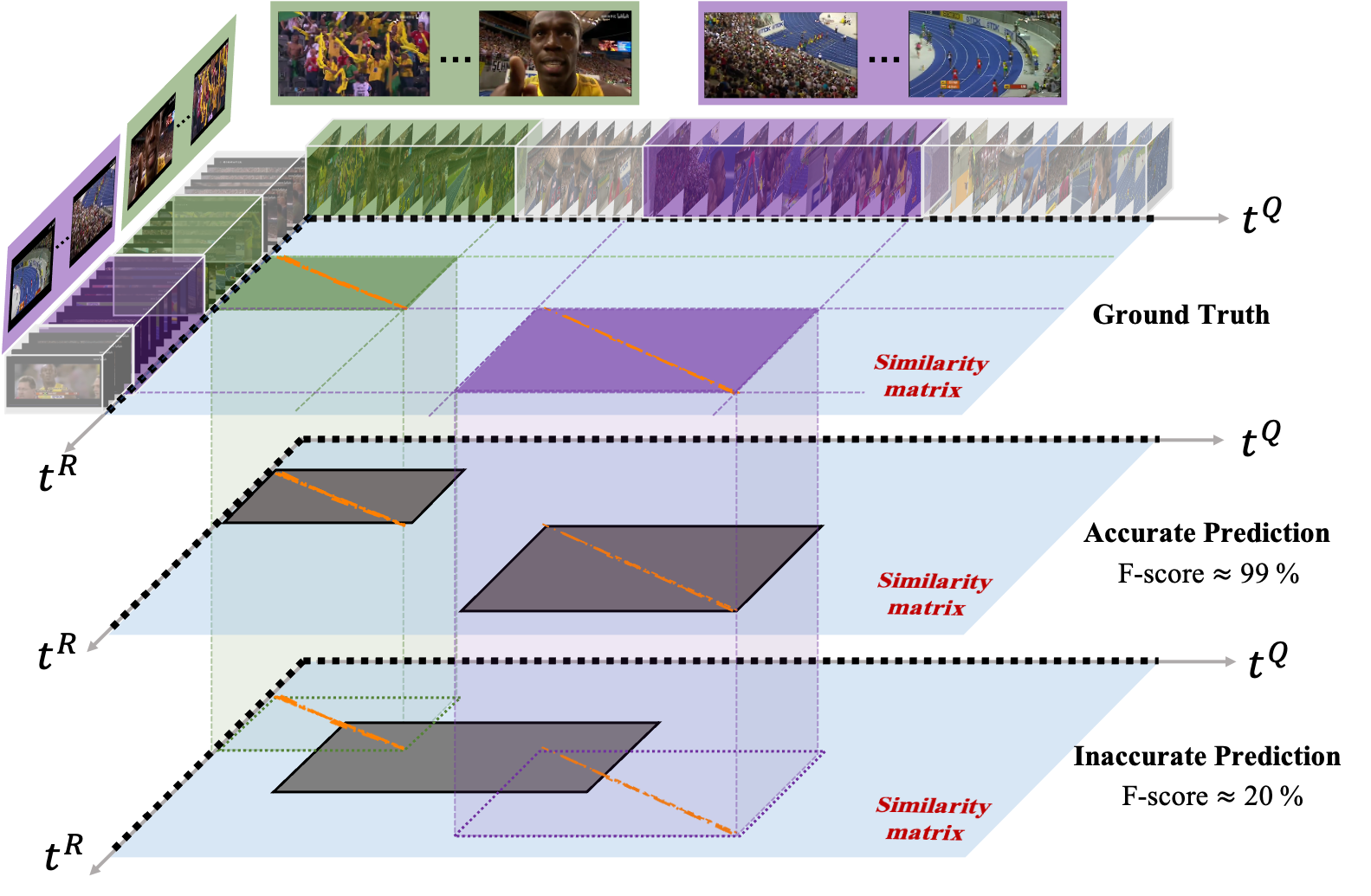} 
\caption{Video copy localization is to accurately detect the start and end {timestamps} of all copied segment {pairs} in the query and reference videos. The yellow highlight points on similarity matrix indicate high similarity scores of temporal corresponding frames between query and reference videos.}
\label{fig1}
\end{figure}

Due to the explosive growth of pirated multimedia and increasing demands for preventing copyright infringements in recent years, content-based video retrieval (CBVR) becomes increasingly essential in applications such as copyright protection, video filtering, recommendation, etc. Besides overall video-level copy detection for given pairs of potential copied videos, finer-grained segment-level copy details with temporal boundaries are also desired particularly in large datasets or real-world scenarios so that applications such as copyright protection become easier and more intuitive. Figure~\ref{fig1} shows this segment-level video copy detection task. The {green} or purple bounding box corresponding to trimmed temporal section of two original videos (query and reference video) is a {pair of} copied segments. The goal of video copy localization task is to search and obtain all these copied segments {pairs} from query and reference videos as consistent as possible to ground-truth labels.

Many recent works ~\cite{lamv,spd, han2021video, he2022large} have been proposed to precisely localize copied segments within video pairs. ~\cite{he2022large} concludes a detailed benchmark pipeline for video copy localization. The pipeline starts with a pair of copied videos as input and then outputs the predicted copied segments. All the benchmarked copy localization algorithms follow the scheme that generates a handcrafted frame-to-frame similarity matrix (usually cosine similarity between frame-level feature sequence pairs), and then feed this correlation matrix to temporal alignment methods including TN ~\cite{temporaNetwork}, DP ~\cite{dp1}, SPD ~\cite{spd} etc. This pipeline has good generalizability and extensibility, and it is suitable for most basic copy transformations.

However, this pipeline also has several limitations.  
First, the frames of a video are commonly processed as individual images, making the modeling of long-range semantic dependencies difficult. 
In addition to the disregard for spatial-temporal relationships {within each video} or between the query and reference videos, the steps of this pipeline are separated and independent. Hence, the input of temporal alignment modules is the similarity correlation which might be severely compressed from initial feature pair. This leads to only local optimization in each module, thus limiting video copy localization performance.


Another important observation is that the data-driven neural networks have significantly outperformed most of conventional methods on video copy detection task \cite{he2022large, visil, spd}. However, the assumption of having a sufficient amount of accurate labeled data for training may not hold, especially for video copy localization task which needs annotations with detailed copy starting and ending timestamps. 
Thus, a natural idea is to leverage abundant unlabeled data or only video-level labeled data to facilitate learning in the original localization task. This requires a large-capacity model that is adaptable to flexible manner of annotations.  

To address the above-mentioned issues, we propose TransVCL, a novel approach to localize copied segments in videos. Motivated by the effectiveness of attention mechanism in capturing temporal dependencies, we propose to incorporate temporal information between {frame-level features} using Transformer \cite{transformer} with self- and cross- attention layers to reason about the internal and mutual relationships between video descriptors, and the similarity matrices are then generated from a differentiable softmax matching layer. After that, the similarity matrices are fed to a temporal alignment network that detects the similarity patterns with detailed starting and ending timestamps under the supervision of labeled copied segments.
Here, the components in our framework including feature enhancement, similarity generation and temporal alignment modules 
are jointly trained to obtain an end-to-end optimization.

Meanwhile, the above all revolve around segment-level copy detection on fine-grained frame correlations, which might lose awareness of global video-level representation. 
Attributed to the architecture of Transformer, we simply add an extra learnable token to the sequence, and this token is to aggregate information from the entire video sequence as a global representation. 
Hence, we supplement an auxiliary video-level binary classification task to further exploit video-level prediction from the relationship between query and reference video representation.
This supplementary global loss is proved to facilitate video copy localization well, while also making better use of video-level labeled data. This enables our algorithm to be more flexibly extended to semi-supervised or weakly supervised learning. 

We evaluate our proposed method on two current segment-level annotated video copy dataset VCSL \cite{he2022large} and VCDB \cite{vcdb}. The experiments show that TransVCL outperforms other algorithms by a large margin with much more accurate copied segment localization. 
Detailed visualization reveals that TransVCL automatically learns to optimize the correlation similarity distribution to a cleaner pattern, and then localize the copied segments on the optimized noiseless similarity matrix.
Moreover, we test TransVCL in semi-supervised setting with using small fraction of labeled dataset and the remainder as unlabeled sets or weakly labeled sets, and TransVCL shows significant accuracy improvements while adding unlabeled or weakly labeled data.
At last, we also utilize publicly available datasets (SVD \cite{svd}, FIVR \cite{fivr}) with only video-level labels to further improve video copy localization performance, and this simulates the real-world applications with continuous streamed unlabeled or weakly labeled data.

\section{Related Work}
\subsection{Video Copy Localization}

Video copy localization, i.e., segment-level video copy detection, is applied to a pair of potential copied videos, including a query video $V^Q=\left \{v_m^Q\right \}_{m=1}^q$ and a reference video $V^R=\left \{v_n^R\right \}_{n=1}^r$, where $v_m^Q$ and $v_n^R$ are $m$-th frame in the query video and $n$-th frame in the reference video respectively, and $q$ and $r$ are the numbers of frames in these two videos. There might exist one or more copied segments between 
$V^Q$ and $V^R$. The objective is to seek the accurate corresponding temporal copied segments $\left \{v_m^Q\right \}_{m=t_{s_1}^Q}^{t_{e_1}^Q}$, $\left \{v_m^Q\right \}_{m=t_{s_2}^Q}^{t_{e_2}^Q}$, ... in $V^Q$ and $\left \{v_n^R\right \}_{n=t_{s_1}^R}^{t_{e_1}^R}$, $\left \{v_n^R\right \}_{n=t_{s_2}^R}^{t_{e_2}^R}$, ... in $V^R$, where $t_{s_i}$ and $t_{e_i}$ are the start and end timestamps and $i$ is the index of multiple copied segments. The frames between $t_{s_i}^Q$ and $t_{e_i}^Q$ in $V^Q$ are the transformed copies of the frames between $t_{s_i}^R$ and $t_{e_i}^R$ in $V^R$. In the segment-level labeled video copy dataset (e.g. VCDB ~\cite{vcdb}, VCSL ~\cite{he2022large}), the detailed copied segment information [$\left \{t_{s_1}^Q, t_{e_1}^Q,t_{s_1}^R, t_{e_1}^R\right \}$, $\left \{t_{s_2}^Q, t_{e_2}^Q,t_{s_2}^R, t_{e_2}^R\right \}$, ...] are provided for full supervision of copy localization. There also exist some video retrieval datasets (e.g. SVD ~\cite{svd}, FIVR ~\cite{fivr}) that only indicate video-level copy information without detailed copied segment localization annotation.

Due to the temporal accuracy requirements on copied segment boundaries, copy localization methods always start with frame-level features rather than clip-level features. 
Meanwhile, for a fair comparison between copy localization algorithms, the frame-level features tend to be fixed. Then a common approach is to generate a frame-to-frame similarity matrix by taking into account of continuous frame sequences, as shown in Figure~\ref{fig1}. In order to obtain the detailed copied localization, a simple method is to vote temporally by temporal Hough Voting (HV) \cite{voting1}. The graph-based Temporal Network (TN) \cite{temporaNetwork} takes matched frames as nodes and similarities between frames as weights of links to construct a network, and the matched clip is the weighted longest path in the network. Another method is Dynamic Programming (DP) \cite{dp1}, whose goal is to find a diagonal block with the largest similarity. Inspired by temporal matching kernel~\cite{Poullot2015}, LAMV\cite{lamv} transforms the kernel into a differentiable layer to conduct temporal alignment. SPD ~\cite{spd} formulates temporal alignment as an object detection task based on the frame-to-frame similarity matrix, achieving a state-of-the-art segment-level copy localization performance.

\subsection{Transformer for Videos }
Transformer ~\cite{transformer} has become the de facto standard for sequence modeling in natural language processing (NLP) due to its simplicity and computation efficiency. 
Recently, Transformers are also getting more attention in basic computer vision tasks, such as image classification ~\cite{vit}, object detection ~\cite{detr} and semantic segmentation ~\cite{segmenter}. 
As for video task, Transformers have also been applied to video retrieval ~\cite{tca}, video super resolution ~\cite{liu2022learning}, video action recognition ~\cite{yang2022recurring}, etc. 
Inspired by the success of Transformers in many vision tasks, we first explore the Transformer architecture for video copy localization task.

{\subsection{Semi-supervised and Weakly-supervised Learning}}
Semi-supervised learning (SSL) exploits the potential of unlabeled data to facilitate model learning with limited annotated data. Most of the existing SSL methods focus on image classification ~\cite{hu2021simple, taherkhani2021self} and object detection ~\cite{sohn2020simple, zhou2021instant} task. On the other hand, the weakly supervised setting, where only global-level category labels are required during training, has drawn increasing attention in image segmentation ~\cite{lee2021bbam,zhou2022regional} and video action localization ~\cite{lee2021weakly, shen2022semi}. In this paper, we introduce these two simple yet effective settings to video copy localization for the first time. Owing to the good adaptability of TransVCL, our approach can be conveniently extended to flexible supervision without requiring complex design.
We believe that the specific network designed only for semi-supervised or weakly supervised video copy localization will also attract increasing attention in the future, but we will not discussed in this paper.

\begin{figure*}[h!]
\centering
\includegraphics[width=\textwidth]{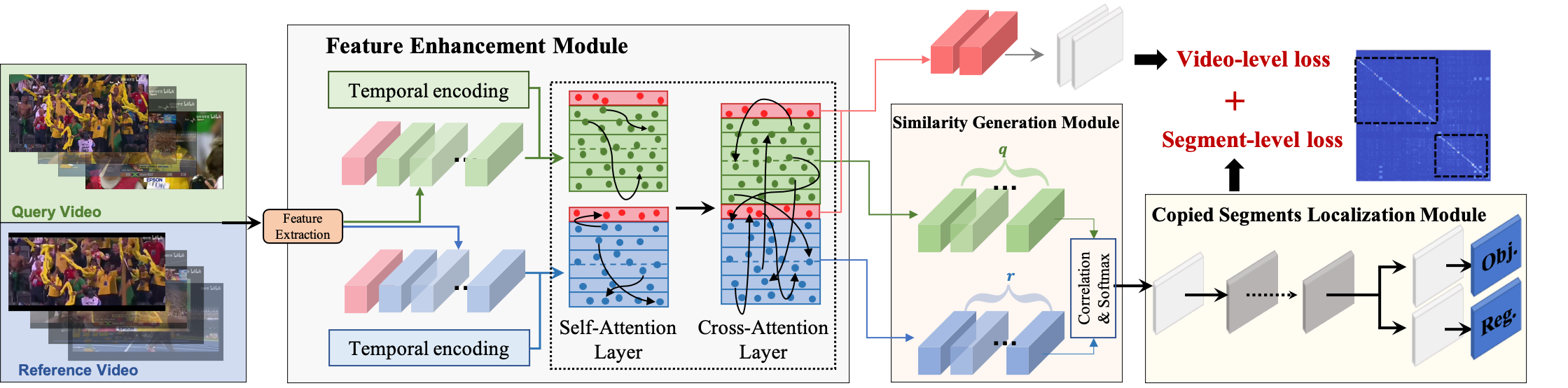} 
\caption{Overview of TransVCL, which is joint trained with three modules including feature enhancement, similarity generation and segment localization. We supplement a video-level loss to facilitate copy localization and further exploit weak labels.}
\label{pipeline}
\end{figure*}

\section{TransVCL Details}
In this section, we introduce the framework details of TransVCL. An overview is presented in Figure \ref{pipeline}, consisting of feature enhancement, similarity generation and segment localization modules. 
Another advantage of flexible extension to semi-supervision, which shows great generalizability of our proposed network, will also be introduced.

\subsection{Feature Enhancement with Attention}
The key to our formulation lies in generating high-quality similarity relationship between the input frame-level feature pairs for next-step alignment. Recall that the frame-level features are extracted independently from a fixed image embedding network. We denote the frame-level feature pair as $F^Q=\left \{f_m^Q\right \}_{m=1}^q$ from query video $V^Q$ and $F^R=\left \{f_n^R\right \}_{n=1}^r$ from reference video $V^R$, and $f_m^Q$ is $m$-th frame-level feature of query video and $f_n^R$ is $n$-th frame-level feature of reference video. To further consider the inter- and intra- dependencies between these features, a natural choice is Transformer, which is particularly suitable for modeling the mutual relationship within or between $F^Q$ and $F^R$ with the attention mechanism. Since the frame-level features have no notion of the temporal information, we first add the fixed sine and cosine temporal encoding to the initial features, which is consistent with positional encoding in ~\cite{transformer}. The sum of initial frame-level feature and temporal encoding also makes the next-step matching process consider not only the feature similarity but also their temporal distance.

Inspired from the leverage of class token in some previous works, like BERT ~\cite{bert},ViT ~\cite{vit}, we prepend a learnable embedding to the sequence of frame-level features on each video, whose value in convergent state serves as the global feature of the given video.
Concretely, the input features to {our }attention modules are {presented as}:
\begin{equation}
\begin{aligned}
  &F_{\mathrm{input}}^Q = [f_{\mathrm{class}}^Q; f_0^Q; f_1^Q; ...] + f_{\mathrm{tem}}\\
  &F_{\mathrm{input}}^R = [f_{\mathrm{class}}^R; f_0^R; f_1^R; ...] + f_{\mathrm{tem}}
  \label{eq1}
\end{aligned}
\end{equation}
where $f_{\mathrm{class}}$ is the learnable {class token} {embedding served} as a global video feature, and $f_{\mathrm{tem}}$ is the temporal encoding.

Then we perform stacked self- and cross- attention modules to {enhance} the initial {input} features. Specifically, for self-attention {module, which considers only the internal relationship inside the frame sequence of a single video}, the query, key and value in the attention mechanism are from the same feature {sequence}, either $F^Q$ or $F^R$. For cross-attention {module}, the key and value are same (e.g., $F^Q$) but different {in} the query (e.g., $F^R$) to introduce their mutual dependencies. This process is performed for both $F^Q$ and $F^R$ symmetrically, which is indicated in Eq.~\ref{eq2}. The formulas of our attention modules are illustrated as below,

\begin{equation}
\begin{aligned}
  &F_{\mathrm{output}}^Q = \mathfrak{T} (F_{\mathrm{input}}^Q, F_{\mathrm{input}}^R)\\
  &F_{\mathrm{output}}^R = \mathfrak{T} (F_{\mathrm{input}}^R, F_{\mathrm{input}}^Q)
  \label{eq2}
\end{aligned}
\end{equation}
where 
\begin{equation}
\begin{aligned}
  \mathfrak{T} &= [\mathfrak{T}_{\mathrm{self}}, \mathfrak{T}_{\mathrm{cross}}] \\
  \mathfrak{T}_{\mathrm{self}}(x, \cdot) &= \mathrm{MHA}(x, x, x) \\
  \mathfrak{T}_{\mathrm{cross}}(x, y) &= \mathrm{MHA}(x, y, y)
  \label{eq3}
\end{aligned}
\end{equation}
and $\mathfrak{T}$ is Transformer, $\mathrm{MHA}$ is multi-head attention {mechanism} with {the same query, key, value} as input in Transformer {but with different focus aspects}.

After attention-based feature enhancement, we can additionally obtain the video-level feature representation for either query or reference videos which is the first token in $F_{\mathrm{output}}^Q$ or $F_{\mathrm{output}}^R$. 
Similar with Next Sentence Prediction in Bert ~\cite{bert}, a binary classification head is attached for further utilization of weak labels (i.e., video-level copy annotations), as well as negative labeled information for uncopied video pairs.
Our classification head is implemented by a MLP with one hidden layer with the concatenation of both class tokens:
\begin{equation}
\begin{aligned}
  y = \mathrm{MLP}([F_{\mathrm{output}}^Q[0]; F_{\mathrm{output}}^R[0]])
  \label{eq4}
\end{aligned}
\end{equation}
where $y$ is the predictive probability of $F^Q$ and $F^R$ being copied in video-level representation. This will be an auxiliary task for copied segments localization (the third modules in TransVCL).

To pursue better efficiency, we propose to use Linear Transformer ~\cite{lineartrans} instead of standard Transformer architecture, an efficient variant of the vanilla attention layer in Transformer, whose complexity is reduced from $O(n^2)$ to $O(n)$ , where $n$ is sequence length. Details of linear attention are attached in Supplementary Material.

{\subsection{Similarity Matrix Generation}}
The frame-to-frame similarity matrix based on frame-level features is capable of representing {similarity relationship} between the copied videos and it has proved to be feasible to search and localize the copied segments 
~\cite{he2022large}. However, aforementioned matching step with generating a similarity map is handcrafted without optimization in learning process. Inspired from local feature matching network ~\cite{rocco2018neighbourhood, loftr}, we apply the differentiable matching layers with a dual-softmax operator to similarity generation. In detail, the similarity matrix is first calculated by correlation expression $S_{mn}= \frac{1}{\tau } \cdot \left \langle \tilde{f}_m^Q, \tilde{f}_n^R \right \rangle $ with a tunable hyper-parameter $\tau$ temperature, and $m$ and $n$ are the temporal frame index of query and reference video respectively, $\tilde{f}$ indicates the attention-enhanced frame-level feature. Then we apply softmax on both dimensions of $S$ to obtain probability of soft mutual nearest neighbor matching:
\begin{equation}
\begin{aligned}
  \tilde{S} = \mathrm{softmax}(S(m, \cdot))_n \cdot \mathrm{softmax}(S(\cdot, n))_m
  \label{eq5}
\end{aligned}
\end{equation}
Then these similarity matrices are reshaped to uniform size, and sent to segment localization module.

\subsection{Copied Segments Localization}

The objective of copy localization task is to find the detailed copied segment information, i.e. [$\left \{t_{s_1}^Q, t_{e_1}^Q,t_{s_1}^R, t_{e_1}^R\right \}$, $\left \{t_{s_2}^Q, t_{e_2}^Q,t_{s_2}^R, t_{e_2}^R\right \}$, ...], 
within the given feature pairs. Interestingly, these 4-dimensional copied segment pairs with start and end timestamps can be formulated as bounding boxes with top-left and bottom-right coordinates in similarity map $\tilde{S}$ 
as shown in Figure \ref{fig1} and Figure \ref{pipeline}. The temporal boundaries of ground truth segment-level labels can also be expressed as bounding box coordinates. 
{Therefore, we treat the task of locating all the copied segment pairs as an object detection task on aforementioned similarity map $\tilde{S}$.} Here, high frame similarities between copied segment pairs compose specific infringement patterns, and this can be learned by the object detection network in a data-driven manner.
The training loss of segment-level supervision is defined as:
\begin{equation}
\label{eq6}
L_{\mathrm{seg}}(F^Q, F^R, p^{\ast}, t^{\ast})=\sum_{i}[L_{obj}(p_i, p_i^{\ast}) + \lambda L_{reg}(t_i, t_i^{\ast})]
\end{equation},
where $i$ is the index of a trimmed copied segment pair detected from input video pair. $p_i$ is the predictive probability of the segment pair being positive (copied). $t_i$ is the 4-dimensional coordinates of the copied segment pair. $p_i^{\ast}$ is the binary label (copied or not) from ground-truth. $t_i^{\ast}$ is temporal location of the ground-truth copied segment pair. We use BCE Loss for training $obj$ branch to predict the trimmed segment pair is similar or not, and IoU Loss for training $reg$ branch to regress their temporal boundaries (coordinates).

In addition, we utilize the obtained video-level copied prediction from Eq.\ref{eq4} for auxiliary binary classification loss:
\begin{equation}
\label{eq7}
L_{\mathrm{video}}(F^Q, F^R, y^{\ast})=L_\mathrm{cls}(y, y^{\ast})
\end{equation},
where $y^{\ast}$ is the ground truth video-level binary label. This weak label can be easily obtained from both segment-level or video-level annotated video copy dataset. The final loss consists of the losses from both segment and video-level: 
\begin{equation}
\label{eq8}
L=L_{\mathrm{seg}} + L_{\mathrm{video}}
\end{equation}

\subsection{Extension to Flexible Supervision}
Due to the network design with both video-level and segment-level task taken into consideration, we can easily adapt TransVCL to semi-supervision and weak supervision. In detail, only part of the provided data of this section are fully labeled with segment-level, and the remainder of data are unlabeled or weakly labeled with only global info (only video-level annotations of copied or not). 

\begin{figure}[htb]
\centering
\includegraphics[width=\linewidth]{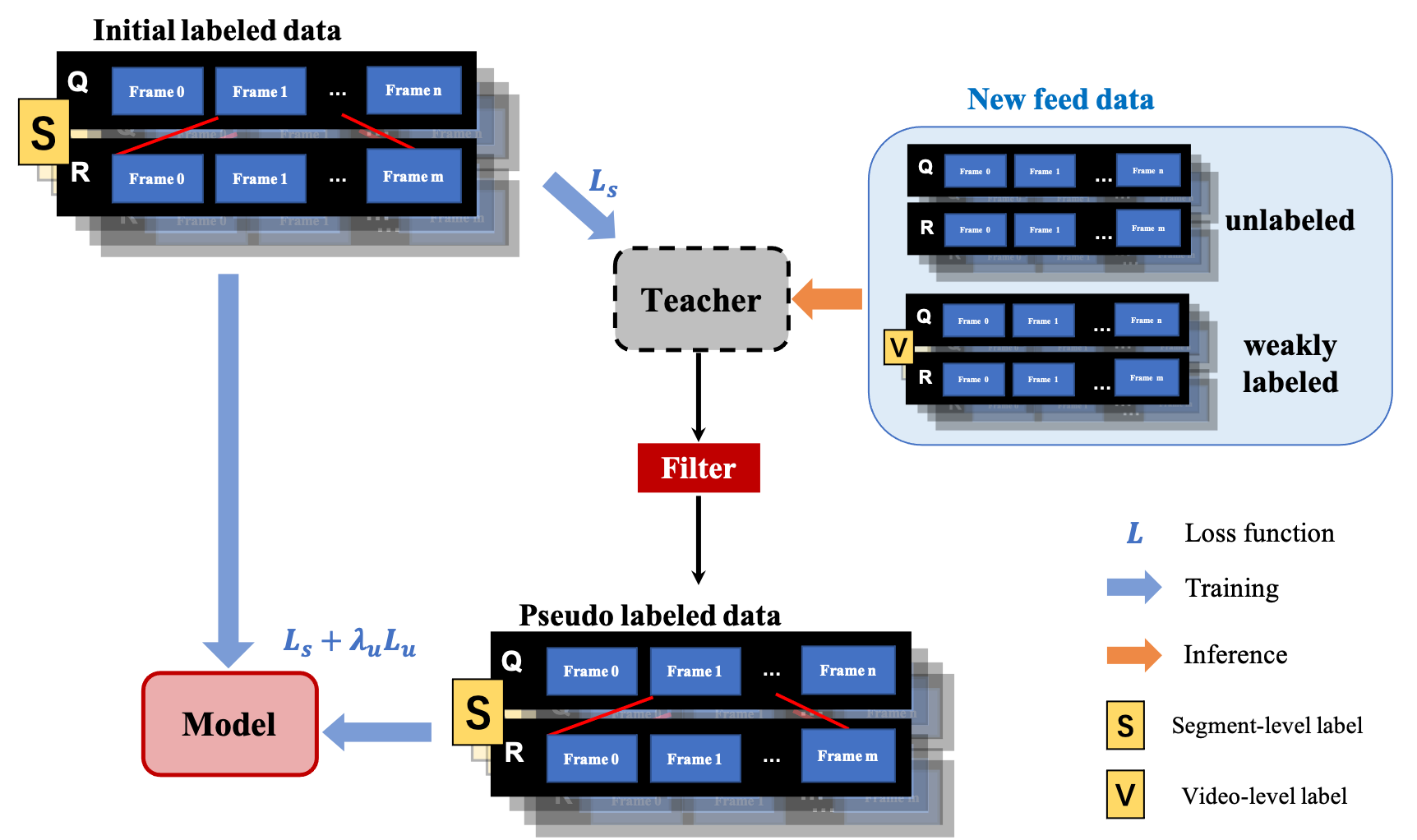} 
\caption{The semi-supervised and weakly semi-supervised setting of video copy localization task.}
\label{fig3}
\end{figure}

Inspired from semi-supervised object detection task ~\cite{sohn2020simple}, a simple framework is proposed to verify the transferability of TransVCL to semi-supervised learning shown in Figure \ref{fig3}. First, we train a teacher model on initial segment-level labeled video feature pairs, and this step is exactly the same as training TransVCL in fully supervised setting above. Second, pseudo labels of unlabeled data and weakly labeled data are generated while inference on the trained teacher model. Third, we use a high threshold value for the confidence-based thresholding to control the quality of pseudo labels comprised of copied segment localization. In this step, the weak supervision with video-level info can also help us filter out the low-quality pseudo labels, and this detailed process will also improve the performance with weakly labeled data. The final step is to train TransVCL network with both labeled and pseudo-labeled data by jointly minimizing the supervised and unsupervised (or weakly supervised) loss as follows:
\begin{equation}
\label{eq9}
\tilde{L} =L_{s}(F^Q, F^R, p^{\ast}, t^{\ast}) + \lambda_{u}L_{u}(F^Q, F^R, y^{\ast})
\end{equation},
where $y^{\ast}$ is the video-level copied labels only provided in weakly supervised situation. Both $L_{s}$ and $L_{u}$ are consistent with Eq.~\ref{eq8}, and $p^{\ast}$ and $t^{\ast}$ in $L_{u}$ come from pseudo labels.

\section{Experiments}
In this section, we first train and evaluate TransVCL on publicly available segment-level video copy datasets with fully supervised setting. The visualization of intermediate results demonstrates the feasibility of our method. The performance of TransVCL to flexible supervision is also provided.

\subsection{Datasets and Evaluation}
{Following the video copy segment localization benchmark ~\cite{he2022large}, we use VCSL dataset for method comparisons.} 
In addition, we also evaluate results on VCDB dataset with smaller scale of data but also containing ground truth copied segments. Besides these two existing datasets with detailed segment-level copied annotations, we further utilize video-level copy dataset (FIVR and SVD) as weakly labeled data for semi-supervised evaluation.

As for evaluation protocol, we take two untrimmed videos as input 
to detect all the potential trimmed copied segments between the two videos. We adopt the F-score, the harmonic mean of segment-level recall and precision ~\cite{he2022large}, as the standard metric. 
This metric not only intuitively indicates segment-level alignment accuracy, but also reflects video-level performance. In addition, coarse-grained video-level false rejection rate (FRR) and false acceptance rate (FAR) are also discussed in ablation study.

\subsection{Implementation Details}
The entire TransVCL network is trained end-to-end with pairs of frame-level video feature sequences extracted from the given query and reference videos as input. 
Due to the unavailability of some raw video files with invalid video links in VCSL dataset, we directly use the official provided ISC features, which are also claimed as SOTA frame-level features in {Facebook AI Image Similarity Challenge} ~\cite{papakipos2022results}. For VCDB and other datasets with raw video files, we {first extract one frame per second and then extract ISC features following \cite{yokoo2021contrastive} like VCSL}, obtaining a 256-dimension embedding for each frame. In the training stage of our experiments, all the video feature sequences are {truncated} with the maximum length of 1200 (i.e., 20min) and padded zeros for sequence length lower than 1200. Therefore, the input to TransVCL network is pairs of video features with uniform size of 1200$\times$256. In the feature enhancement component, the head of Transformer is 8.
In the similarity generation component, temperature $\tau$ of dual-softmax is 0.1 and the similarity matrix is reshaped to (640, 640). In copied segment localization module, we adopt the anchor-free detection network YOLOX ~\cite{ge2021yolox} with simple design, and the regression loss weight $\lambda$ is 5 as default. The entire model is trained using SGD with momentum 0.9, batch size of 64, initial learning rate of 0.01 and weight decay of 0.0005. The detailed implementation can also be inferred in our codes.

{\subsection{Comparison with Previous Methods}}
We compare our method to the previously reported copy localization algorithms in VCSL benchmark \cite{he2022large}. In addition, we also evaluation these methods on VCDB dataset as shown in Table \ref{table1}. All these methods are trained (if necessary) on close to 100k pairs of copied video pairs and an approximately equal number of automatically generated negative uncopied pairs between different video categories in train set of VCSL dataset. As can be shown from Table \ref{table1}, TransVCL outperforms all other methods by a large margin on VCSL dataset, with F-score of 66.51\% (+4.55\% compared with previous SOTA SPD). On VCDB dataset, TransVCL also achieves the best F-score of 75.37\%. Based on the attention-enhanced features and end-to-end optimization, our proposed method obtains the highest accuracy on copy segments localization. 
In detail, TransVCL achieves the best results on over 60\% of the topic categories (7/11) including the most common daily life videos and the most difficult kichiku videos. Meanwhile, 
TransVCL shows significantly higher accuracy ($+10\%$ higher F-score than second-best) on short copied segments within videos (copy duration percentages $<$ 40\%). For other topic categories and copy percentages scenarios, TransVCL mostly achieves top-2 highest performances which are very close (only $0.5\sim1\%$ F-score) to the highest scores. If we look into some individual hard cases, TransVCL also presents considerably better localization results on some complex video transformations including backwards-running, acceleration or deceleration, short copied segments, mushup, etc. The detailed statistic comparison and some visualization cases are given in Supplementary Material.

\begin{table}[h!]
\caption{Video copy localization performance comparison of our proposed TransVCL with previous benchmarked approaches on VCSL and VCDB datasets. }
\centering
\label{table1}
\scalebox{1}{
\begin{tabular}{c|l||c|c|c}
\toprule
{Dataset} & {Method}   & Recall & Precision & \textbf{F-score$\uparrow$} \\ \toprule
\multirow{6}{*}{\begin{tabular}[c]{@{}c@{}}VCSL\end{tabular}}    & HV        & 86.94  & 36.82     & 51.73   \\  
                                                                             & TN        & 75.25  & 51.80     & 61.36   \\  
                                                                             & DP        & 49.48  & 60.61     & 54.48    \\ 
                                                                             & DTW       & 45.10  & 56.67     & 50.23    \\ 
                                                                             & SPD       & 56.49  & 68.60     & 61.96    \\ 
                                                                             & TransVCL & 65.59  & 67.46     & \textbf{66.51}   \\ 
                                        \midrule                                 
\multirow{6}{*}{\begin{tabular}[c]{@{}c@{}}VCDB\end{tabular}} & HV        & 89.23  & 58.70     & 70.81   \\ 
                                                                             & TN        & 80.06  & 69.20     & 74.23   \\  
                                                                             & DP        & 63.84  & 73.54     & 68.35   \\  
                                                                             & DTW       & 61.78  & 72.26     & 66.61   \\ 
                                                                             & SPD       & 71.00  & 78.82     & 74.71    \\ 
                                                                             & TransVCL & 76.69  & 74.09     & \textbf{75.37}   \\  
                                                                         \bottomrule
\end{tabular}}
\end{table}

\subsection{Ablation Study}

\begin{table}[]
    \centering
    \caption{Ablation of detailed settings in TransVCL network.}
    \label{table2}
    \begin{tabular}{l||c|c|c}
        \toprule
        {Method} & Recall & Precision & \textbf{F-score$\uparrow$}  \\ \toprule
        {basic localization model} & 60.24  & 64.70 & 62.39\\ 
        { +self-att\&cross-att} & 63.96 & 65.54 & 64.74 \\ 
        { +att+video-level loss} & 60.19  & 71.43 & 65.33 \\ 
        { +att+temporal encoding} & 65.88  & 66.79 & 66.34 \\ 
        {full model} & 65.59  & 67.46 & {66.51} \\  \bottomrule
    \end{tabular}
\end{table}

We ablate different settings of feature enhancement module on VCSL dataset in Table \ref{table2}. {Here, the basic localization model has the same setting as the aforementioned copied segment localization module.}
Compared with the basic localization model with conventional frame-to-frame cosine similarity map as input, the F-score is significantly improved after incorporating the attention module and joint optimization. 
Meanwhile, the class token along with the auxiliary branch of video-level classification loss introduces a broader guidance to model training, bringing additional performance gains. Temporal encoding component embeds the sequential positional information to video features, which also improves the copy localization performance.

In Table \ref{table2}, the global class token design as an auxiliary task is proved to facilitate the copied segment localization accuracy. To further demonstrate the feasibility of our design, we also show coarse-grained copy detection results with video-level FRR/FAR in Table \ref{table3}. The video-level performance of basic localization model slightly improves compared with previous SOTA SPD algorithm, and the error rate is significantly lower after joint optimization with our designed attention module. Meanwhile, the video-level result of our method is also better than any others reported in video copy segment localization benchmark ~\cite{he2022large}.

\begin{table}[]
    \centering
    \caption{Video-level copy detection performance gain from the attention module.}
    \label{table3}
    \begin{tabular}{l||c|c|c}
        \toprule
        {Method} & {FRR $\downarrow$}  &{FAR $\downarrow$} & \textbf{F-score $\uparrow$}  \\ \toprule
        {SPD} & 0.2974 & 0.0958 & 79.08 \\ 
        {basic localization model} & 0.2527  & 0.1105 & 81.22\\ 
        {full model} & 0.1666  & 0.0173 & {90.19} \\  \bottomrule
    \end{tabular}
\end{table}

\subsection{Visualization of Intermediate Results}

\begin{figure*}[ht!]
\centering
\includegraphics[width=1\textwidth]{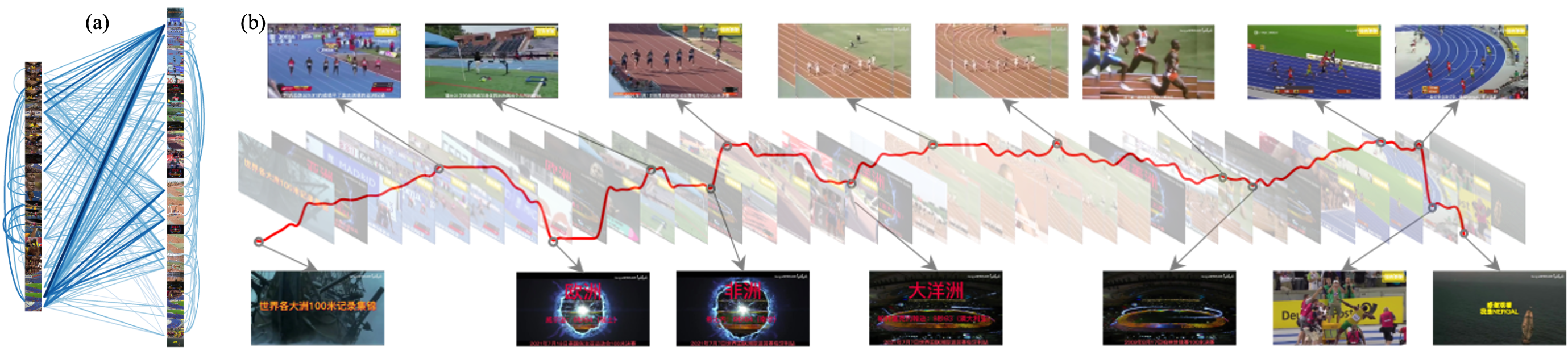} 
\caption{(a)Visualization of attention weights. The left frames are from query video and right frames are from reference video. The thickness of lines indicate strength of attention weights. Zoom in for details. (b)Attention weights {about} the global feature (class token) to frame feature sequence. After training, local maximum clips of attention (red line) are learned to be the highlight moments of reference videos (top row), and local minimum points are mostly unimportant transitions (bottom row).}
\label{fig4}
\end{figure*}

\begin{figure}[ht!]
\centering
\includegraphics[width=0.92\linewidth]{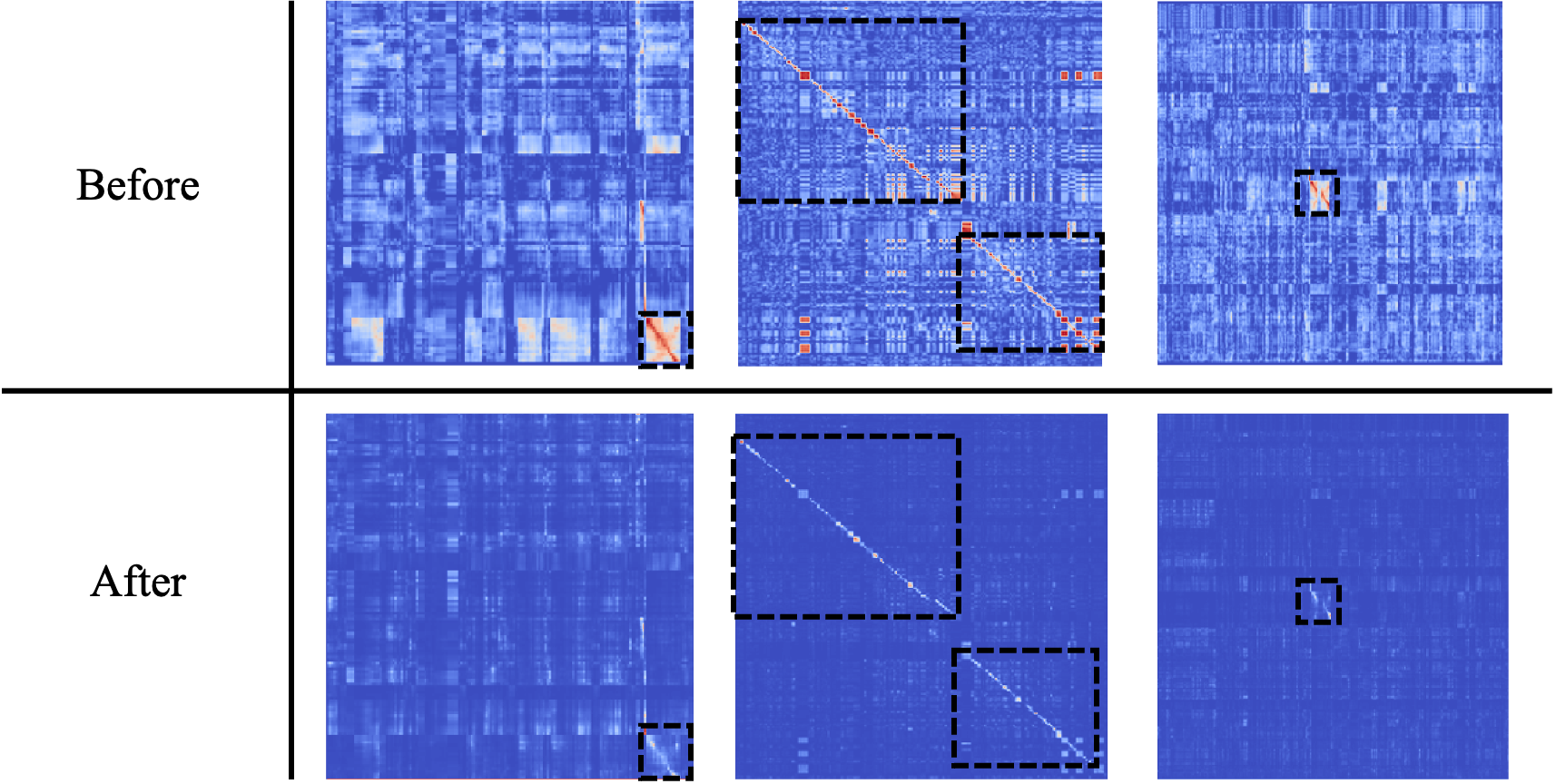} 
\caption{Learned intermediate results, i.e., similarity matrices, before (upper row) and after (bottom row) attention-based feature enhancement for three example copied video pairs. The similarity patterns in black dashed rectangle are the ground-truth copied segment localization.}
\label{map}
\end{figure}

To better understand the effect of our proposed method, we visualize self- and cross- attention weights in Figure~\ref{fig4}(a). The self-attention represented as arcs establishes long-range semantic connections within the video, {thus expanding the receptive field from individual frames or clips to almost the entire video}. Meanwhile, the cross-attention enhances the correlations between the input video pair. For example, the semantic correlation of frames such as {the relevant screenshot of runway in} {Bolt's} 100-meter race is established even with long temporal interval in Figure~\ref{fig4}(a). As a result, informative frames describing key and relevant moments of the event get higher response, and the redundant frames are suppressed, leading to significantly cleaner frame-to-frame similarity maps shown in Figure~\ref{map}. This learned intermediate result reduces the difficulty of follow-up copied segment localization and obtains obvious performance gains.

Moreover, we also observe how global feature integrates information across the video sequence and further improves the video copy detection performance. The red curve in Figure~\ref{fig4}(b) presents the attention weights between the class token and different frame features in reference video. Here, query video is the same Bolt's 100-meter race video in Figure~\ref{fig4}(a). It is interesting that the global feature's attention learns to peak at the temporal location of video highlights, i.e., clips of 100-meter race scene. This learned mechanism also enhances the effective information and reduces noisy disturbance, bringing a more robust video copy detection accuracy. In addition, the visualization in Figure~\ref{fig4}(b) also shows the potential of our method for extending to other video tasks such as video summarization or event retrieval.


{\subsection{Performance in Semi-supervised Setting}}

\begin{table*}[h!]
    \centering
    \caption{Comparison in F-score for different semi-supervised settings on VCSL dataset. 1\%, 2\%, 5\% and 10\% of labeled training data are randomly sampled as a labeled set (with detailed copied segments locations) and we use the rest of labeled training data as an unlabeled or weakly labeled (only video-level copy annotation) set. }
    \label{table4}
    \begin{tabular}{l||c|c|c|c|c}
        \toprule
        {Method} & 1\%VCSL & 2\%VCSL & 5\%VCSL & 10\%VCSL & 100\%VCSL  \\ \toprule
        {Supervised on labeled data only} & 23.10  & 31.95 & 47.71 & 59.07 & 66.51\\ 
        {Semi-supervised (w/o weak label)} & 37.28 (+14.18) & 44.80 (+12.85) & 56.11 (+  8.40) & 61.98 (+2.91)& / \\ 
      {Semi-supervised (with weak label)} & 44.87 (+21.77) & 50.50 (+18.55) & 58.03 (+10.32)& 62.60 (+3.53) & / \\  \bottomrule
    \end{tabular}
\end{table*}

\begin{table}[]
    \centering
    \caption{Cross dataset semi-supervised learning performance of supplementing video-level labeled datasets.}
    \label{table5}
    \begin{tabular}{l||c|c}
        \toprule
        {Method}  & {F-score} & $\Delta$  \\ \toprule
        {Supervised on 10\% VCSL} & 59.07  & /\\ 
        { +FIVR \& SVD (weak label)} & 63.04 & +3.96  \\ \hline
        {Supervised on 100\% VCSL} & 66.51  & /\\ 
        { +FIVR \& SVD (weak label)} & 67.13 & +0.62  \\ 
 \bottomrule
    \end{tabular}
\end{table}

Our proposed method can also be flexibly extended to semi-supervised learning, and we experiment two settings here. The first is to randomly sample 1\%, 2\%, 5\% and 10\% of labeled training data in VCSL and use rest of training data as an unlabeled or weakly labeled set. The 1\%, 2\%, 5\% and 10\% protocol also means that the scale ratio between unlabeled (or weakly labeled) data and labeled data is 99$\times$, 49$\times$, 19$\times$ and 9$\times$. Besides sampling within a single dataset, we also use the entire VCSL as a labeled set and additional video-level labeled dataset (FIVR, SVD) as a weakly labeled set. In detail, FIVR and SVD provide $\sim17.8$k weakly labeled copied video pairs, which are corresponding to 1.8$\times$ and 0.18$\times$ scale of 10\%VCSL and 100\%VCSL ($\sim97.7$k) respectively. This second protocol evaluates the potential to improve our SOTA video copy localization method with weakly labeled data.
In addition, semi-supervision also introduces another hyper-parameters $\lambda_u$ as the unsupervised loss weight in Eq. 9. After studying the impact of $\lambda_u$ on 10\% protocol (details in Supplementary Material), the best performance 
is obtained when $\lambda_u = 0.5$. 
Two semi-supervised protocol results are summarized in Table \ref{table4} and Table \ref{table5}. 

For 1\%, 2\%, 5\% and 10\% protocols, we demonstrate that TransVCL can be effectively extended to semi-supervised learning with significant F-score improvements from unlabeled data, especially under high ratio between unlabeled and labeled data. Meanwhile, the semi-supervised results are further improved from unlabeled data to weakly labeled data, bringing a considerable performance gain with controllable annotation labor. This improvement is also obvious when the scale of unlabeled data is much larger than labeled data, and this is extremely suitable for online streamed data in practical use. From the cross dataset performance in Table \ref{table5}, F-score is also improved and the SOTA performance is further refreshed from 66.51 to 67.13 while only adding 0.18$\times$ scale of more data with weak label. Notably, on the same 10\% labeled data, FIVR \& SVD brings slightly higher improvements (+3.96) than 90\% VCSL (+3.53), even though the weakly labeled data of former is much less than latter (1.8$\times$ vs 9$\times$). Our proposed network is not only suitable for weakly supervised scenarios, but also has a good capacity to the data diversity. The detailed precision and recall results are given in Supplementary Material.

\section{Conclusion}
We have presented a novel network named TransVCL with joint optimization of multiple components for segment-level video copy detection and demonstrated its strong performance. Enhanced by attention mechanism, TransVCL incorporates temporal correlation within and across input videos and captures more accurate copied segments. Meanwhile, the flexible extension to semi-supervised and weakly semi-supervised scenarios further demonstrates the advantages of TransVCL. We hope our new perspective will pave a way towards a new paradigm for accurate and data-efficient video copy localization.

\section{Acknowledgments}
    This work is partly supported by R\&D Program of DCI Technology and Application Joint Laboratory.

\bibliography{aaai23}

\end{document}


\maketitle
\section{Linear Transformer}
In Section \textit{Feature Enhancement with Attention} of the main text, we propose to use Linear Transformer instead of standard Transformer architecture to pursue better efficiency. In this Section, we will introduce the details of Linear Transformer.

\begin{figure}[t]
\centering
\includegraphics[width=1.0\linewidth]{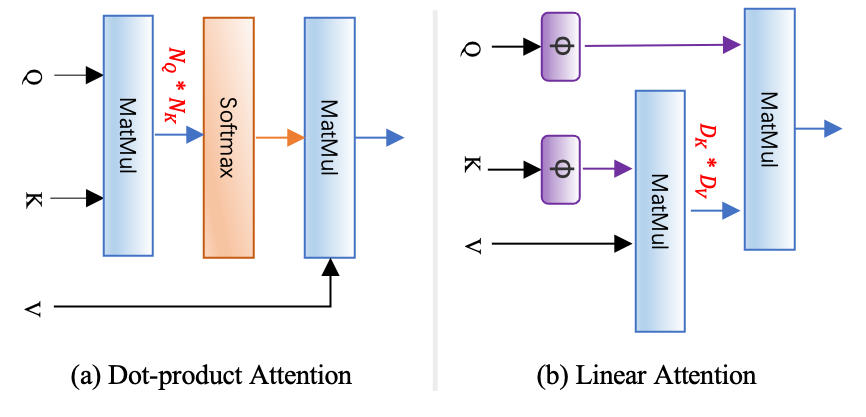} 
\caption{Attention kernel of vanilla transformer and linear transformer structure. }
\label{sup1}
\end{figure}

Transformer has a powerful ability in sequence modeling and thus it is widely used in natural language processing(NLP). It has also been applied to computer vision tasks with input of images as patch sequences or videos as frame sequences. The key element where Transformer contributes to understand long-range information in given sequence is the attention layer. It takes vectors named query, key and value as inputs and retrieves the query vector $Q$ information from the value vector $V$, according to the attention weight computed from the dot product of $Q$ and the key vector $K$ corresponding to each value $V$. This leads to quadratic complexity of the input length. The attention layer is denoted formally as:
\begin{equation}
\begin{aligned}
  Attention(Q, K, V)  = softmax(QK^{T})V.
  \label{eq1}
\end{aligned}
\end{equation}
Usually, Transformer is composed of multiple heads of attention, which leads to a higher computation cost. Since the complexity becomes a knotty problem in practical application with streamed data to take in, it demands handling. An efficient variant of vanilla Transformer named Linear Transformer is proposed to remedy the computation problem and reduce the complexity to $O(n)$. The original attention layer shown in Figure \ref{sup1}(a) uses dot-product to function on $Q$ and $K$ and the computation cost is $N_{Q}*N_{K}$ which makes it exponential. Linear Transformer shown in Figure \ref{sup1}(b) substitutes it to a linear operation sim$(Q, K)=\phi(Q)\cdot\phi(K)^{T}$, where $\phi(\cdot) = elu(\cdot) + 1$. This function reduces the computation to $D_{k}*D_{v}$, where $D$ is the feature dimension and is far smaller than $N$.
As our task is to localize video copy segments in each given video pair on a large dataset, the efficiency is important, thus it is beneficial to adopt the attention layer of the linear variant of Transformer rather than the vanilla version.

\begin{figure}[t]
\centering
\includegraphics[width=1.0\linewidth]{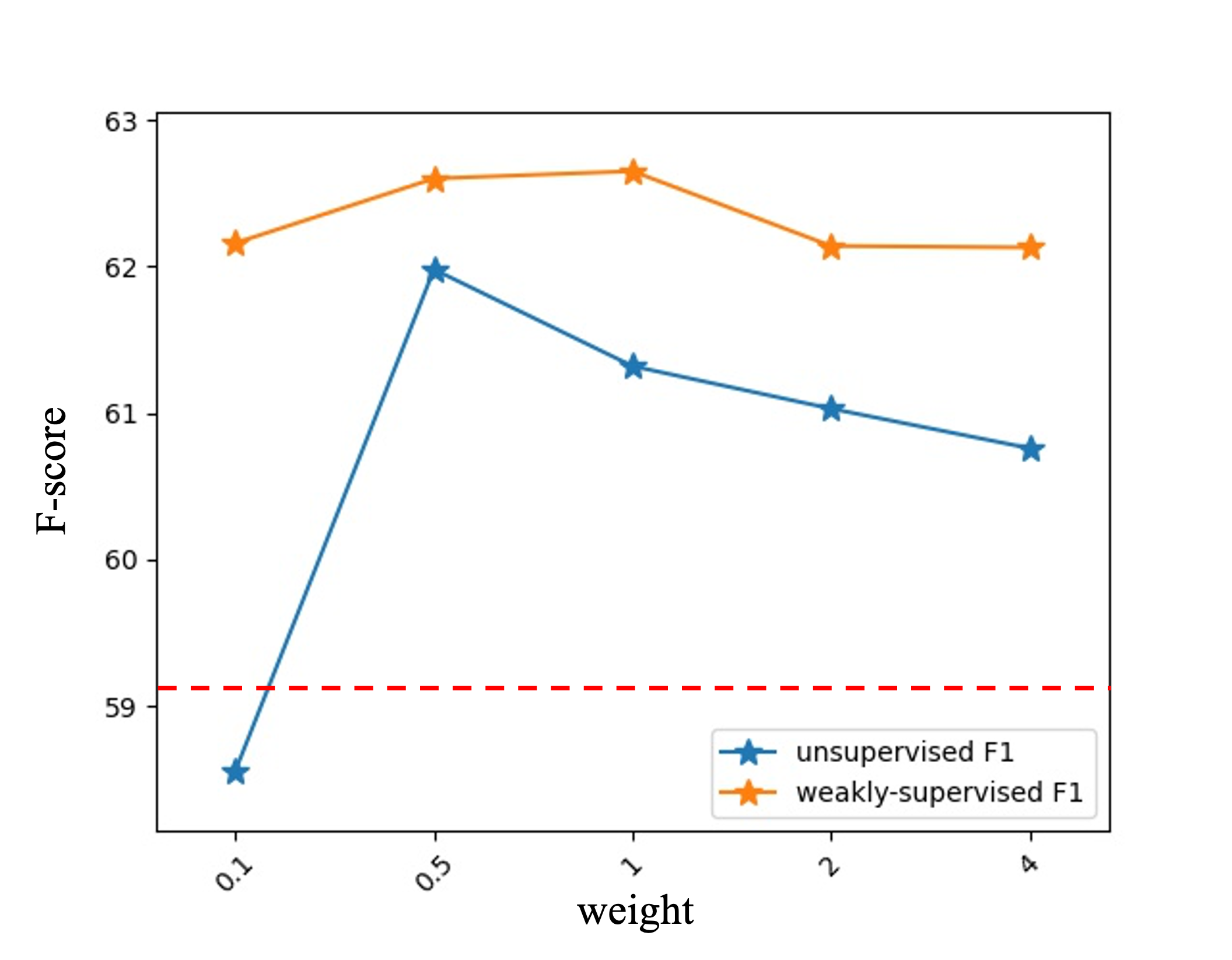} 
\caption{F-score curve to show how the F-score results change with the weight for the unsupervised loss.}
\label{F1}
\end{figure}


\begin{table*}[]
    \centering
    \caption{F-score results of different methods at different topic categories. \textbf{\color{RoyalBlue}{Best}} and \textbf{\color{CornflowerBlue}{second best}} results are highlighted.}
    \begin{tabular}{l||c|c|c|c|c|c|c|c|c|c|c|c}
        \toprule
        {Topic/} & Variety & Games & Music & News & Sports & Life & Kichiku & Advertise & Anima & Films & TV \\
        {Methods}&  Show &  &  &  &  & &  & -ments & -tions &  & series \\ \hline
        {HV} & 83.08 &94.40 &71.99 &83.91 &37.27&52.59 &53.77 &63.13 &76.99 &44.86 &88.40\\
        {TN} & \textbf{\color{RoyalBlue}{97.03}} &\textbf{\color{RoyalBlue}{96.44}} &78.74 &88.11 &55.23 &51.19 &\textbf{\color{CornflowerBlue}{58.95}} &\textbf{\color{CornflowerBlue}{72.77}} &91.68 &62.23 &\textbf{\color{RoyalBlue}{95.13}}\\
        {DP} & 83.27 &92.46 &72.48 &87.76 &51.58 &37.78 &47.34 &68.97 &87.32 &59.27 &88.70\\
        {DTW} & 78.43 &90.72 &68.69 &81.19 &42.79 &46.46 &49.87 &59.69 &91.45 &47.75 &83.24\\
        {SPD} & \textbf{\color{CornflowerBlue}{96.39}} &95.06 &\textbf{\color{RoyalBlue}{89.56}} &\textbf{\color{CornflowerBlue}{89.24}} &\textbf{\color{CornflowerBlue}{61.92}} &\textbf{\color{CornflowerBlue}{52.78}} &46.62 &66.06 &\textbf{\color{CornflowerBlue}{95.79}} &\textbf{\color{CornflowerBlue}{62.76}} &92.99\\
        {transVCL} 
        & 94.89 
        &\textbf{\color{CornflowerBlue}{96.06}} &\textbf{\color{CornflowerBlue}{86.53}} &\textbf{\color{RoyalBlue}{91.23}} &\textbf{\color{RoyalBlue}{67.43}} &\textbf{\color{RoyalBlue}{66.44}} &\textbf{\color{RoyalBlue}{59.21}} &\textbf{\color{RoyalBlue}{74.18}} &\textbf{\color{RoyalBlue}{96.08}}
        &\textbf{\color{RoyalBlue}{72.64}} 
        &\textbf{\color{CornflowerBlue}{93.54}}\\
        \hline
    \end{tabular}
    \label{tab:topic}%
\end{table*}

\begin{table*}[]
\centering
  \caption{Comparison in Recall / Precision / F-score for different semi-supervised settings on VCSL.}
  \resizebox{\textwidth}{!}{\begin{tabular}{c||c|c|c|c}
   \toprule
    {Methods} & 1\% & 2\%  & 5\% & 10\% \\
    \toprule
    {Supervised on labeled data only}& 15.08 / 49.36 / 23.10 & 48.06 / 23.92 / 31.95 & 44.56 / 51.34 / 47.71 & 57.97 / 60.22 / 59.07 \\
    \hline
    {Semi-supervised (w/o weak label)} & 50.52 / 29.54 / 37.28 & 41.03 / 49.34 / 44.80 & 57.16 / 55.11 / 56.11 & 64.91 / 59.30 / 61.98\\
    \hline
    {Semi-supervised (with weak label)} & 40.72 / 49.97 / 44.87 & 51.72 / 49.34 / 50.50 & 52.34 / 65.12 / 58.03 & 66.69 / 58.98 / 62.60\\
    \bottomrule
  \end{tabular}}%
  \label{tab:percent}%
\end{table*}

\begin{table}[]
    \centering
    \caption{F-score results of different methods at different copy duration percentages. \textbf{\color{RoyalBlue}{Best}} and \textbf{\color{CornflowerBlue}{second best}} results are highlighted.}
    \resizebox{\linewidth}{!}{\begin{tabular}{l||c|c|c|c|c}
        \toprule
        {Ratios} & 0-20\% & 20-40\% &40-60\%&   60-80\%&    80-100\% \\  \hline
        {HV} & 15.07 &36.12 &55.98 &68.84 &85.38\\
        {TN} & 39.36 
        &\textbf{\color{CornflowerBlue}{52.60}} &\textbf{\color{RoyalBlue}{69.55}} &\textbf{\color{RoyalBlue}{72.88}} &80.93\\
        {DP} & 39.85 &51.59 &58.32 &58.52 &73.19\\
        {DTW} & 25.85 &43.78 &50.46 &60.46 &80.78\\
        {SPD} & \textbf{\color{CornflowerBlue}{47.97}} 
        &50.07 &62.57 &58.51 &\textbf{\color{CornflowerBlue}{82.25}}\\
        {TransVCL} &\textbf{\color{RoyalBlue}{58.87}} &\textbf{\color{RoyalBlue}{62.53}} &\textbf{\color{CornflowerBlue}{68.57}} &\textbf{\color{CornflowerBlue}{72.34}} &\textbf{\color{RoyalBlue}{87.62}}\\
         \hline
    \end{tabular}}
    \label{tab:duration}%
\end{table}

\begin{table}[]
    \centering
    \caption{Comparison in Recall, Precision and F-score for cross dataset semi-supervised learning.}
    \label{table5}
    \resizebox{\linewidth}{!}{\begin{tabular}{l||c|c|c}
        \toprule
        {Method} & Recall & Precision & F-score  \\ \toprule
        {Supervised on 10\% VCSL} & 57.97 &  60.22 & 59.07\\ 
        { +FIVR \& SVD (weak label)} & 66.89 & 59.62 & \textbf{63.04}  \\ \hline
        {Supervised on 100\% VCSL} & 65.59 & 67.46 & 66.51 \\ 
        { +FIVR \& SVD (weak label)} & 69.66 & 64.77 & \textbf{67.13}   \\ 
    \bottomrule
    \end{tabular}}
    \label{tab:cross}%
\end{table}

\section{Details of Comparison with Previous Methods}
We compare our method to the previously reported copy localization algorithms in VCSL benchmark in our main text. In this part of Supplementary Material, more fine-grained results of at different topic categories and at different copy duration percents are shown in Table~\ref{tab:topic} and Table~\ref{tab:duration}, with blue and bold text highlighted on the best two F-score performance among all the temporal aligned methods. 
We use royal blue to highlight the best performance, and use cornflower blue to highlight the second best. We can conclude from Table~\ref{tab:topic} that our method outperforms other methods in most topic categories, which achieves best performance in seven topics and second best in three topics. As we divide the video dataset into eleven topic categories, we achieve best performance at a ratio of 7 to 11 (over 60\%). For News, Sports, Life, Kichiku, Advertisements, animations and Films, our method win the best score. And for Games, Music and TV series, our results are the second best. The only topic category that our method fails to achieve best two scores is variety show, but still achieves good performance inferior to TN and SPD with a little behind. Our method outperforms the second best at Life and Films topic categories by a large margin, which is about +10\%.  
Moreover, SPD and TN also show some superiority in part of aspects. TN achieves 5 top-2 scores and three highest scores in variety shows, games and TV series, and SPD achieves 7 top-2 best and one highest in music. 

We also evaluate the algorithms at different copy duration percentage shown in Table~\ref{tab:duration}. We take the average length of the query and reference videos, and calculate the percentage between the total duration of the ground-truth copied segments and the average video length. We take 20\% as an interval to calculate the F-score results obtained by different methods. Among all the duration percent intervals, our method achieves the highest F-score on more than half (3/5) and obtains second highest in the rest duration percent intervals. To be more specific, our method has three highest scores in 0-20\%, 20-40\% and 80-100\% and the scores of these three percent intervals are much higher than the second highest. For the percent intervals of 40-60\% and 60-80\%, in which we achieve second high scores, are very close to the highest scores that TN achieves. The results in Table~\ref{tab:topic} and Table~\ref{tab:duration} show robustness of TransVCL clearly.

\section{Detailed Semi-supervision Results}
We introduce an important hyper-parameter $\lambda_u$ for the unsupervised loss in Eq.9 of main text during semi-supervised training. To visualize how the $\lambda_u$ affects the result, we present a F-score curve in Figure~\ref{F1} to show how the F-score results change with the weight under the two semi-supervised learning settings with/without video-level annotations. We choose 0.1, 0.5, 1, 2, 4 for weight in this experiment. By analyzing the F-score curve, we find that the best overall performance of semi-supervised without and with weakly labeled is obtained when $\lambda_u = 0.5$. The F-score are 61.98\% and 62.60\% respectively. For the comparative protocol with only 10\% of VCSL training data as supervised data, our method get a F-score of 59.07, which is a baseline to two polygonal lines in Figure~\ref{F1}, and it is presented as a red dotted horizontal line. 
Our method obtains improvements with feeding in rest data of VCSL without any extra annotation shown as blue curve. And the results are further improved while the rest data of VCSL contain video-level annotations shown as orange curve. 

In addition to overall F-score performance which is already presented in section \textit{Performance in Semi-supervised Setting} of the main text, we also provide more details about Recall/Precision results in Table~\ref{tab:percent} and Table~\ref{tab:cross} (corresponding to Table 4 and Table 5 in our main text). As mentioned in main text, our proposed method can be flexibly extended to semi-supervised and weak-supervised learning, and we experiment two settings. The first is to randomly sample 1\%, 2\%, 5\% and 10\% of label training data of VCS and use rest of training data as an unlabeled or weakly labeled set. The 1\%, 2\%, 5\%, 10\% protocol means the scale ratio of unlabeled (or weakly labeled) data to labeled data is 99$\times$, 49$\times$, 19$\times$, 9$\times$. The detailed results with precision and recall are shown in Table~\ref{tab:percent}. The second is to add additional video-level labeled dataset (FIVR, SVD) as a weakly labeled set to a already large-scale dataset (VCSL), improving performance further as shown in Table ~\ref{tab:cross}. The details of the setting are illustrated in the main text and please refer to that if needed.

\section{Visualization of localization results }
In this Section, we visualize the video copy localization results of different methods on some hard cases. These hard cases of video transformations include short copied segments within similar videos, multiple short copied segments, dramatic \textit{picture in picture} frame transformation, backwards-running videos, acceleration or deceleration speed of videos, and we show these 5 typical cases in Figure 3$\sim$6 respectively. In each figure, we first give the uniformly sampled frames from the query video (vertical) and the reference video (horizontal) and plot the similarity matrices between this pair of videos in the corresponding positions with color map. The black dashed boxes on the color similarity matrices are the ground truth copied segments localization. Moreover, we also provide the visualized copy detection results of different benchmarked methods. The localization results are also plotted on the similarity map but with red boxes. In order to show the localization performance more clearly, the similarity map of different methods are shown in gray-scale (the content information is exactly the same as color similarity map only with different appearance). We also denote the F-score of different methods on these figures. It can be observed that our proposed method (TransVCL) presents considerably better results on these complex video copy transformations. The detailed analysis is given in the figure captions of Figure 3$\sim$6.

\begin{figure}[ht!] 
  \centering
  \includegraphics[width=1\linewidth]{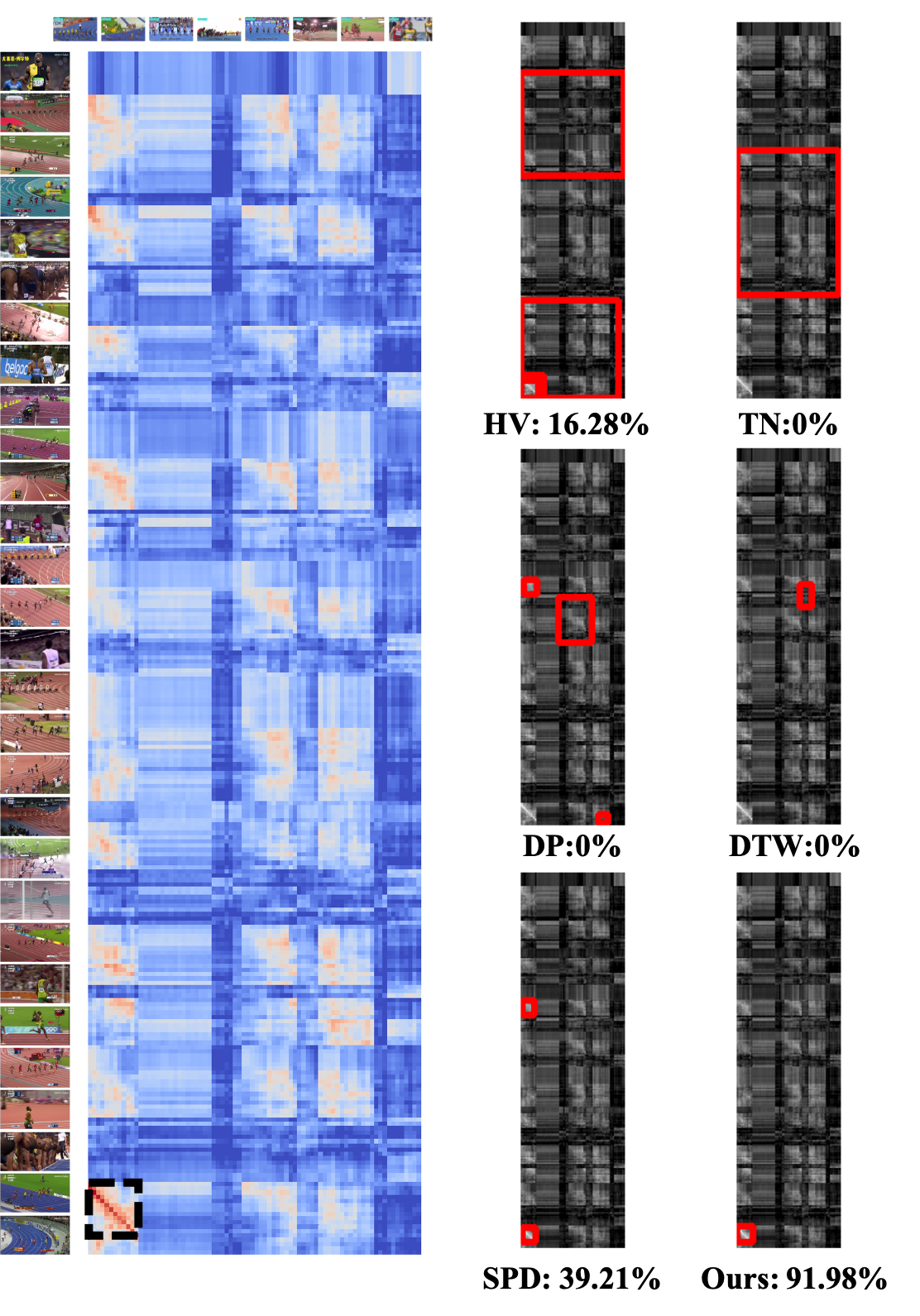}
  \caption{Example of video copy localization results. This example shows a hard case of short copied segments within similar videos. Only our approach can accurately localize the copied part (Bolt's world record of 9.58 seconds) in each video, without producing false positives on other similar but not copied video segments.}
  \label{sim}
\end{figure}

\begin{figure}[ht!] 
  \centering
  \includegraphics[width=0.8\linewidth]{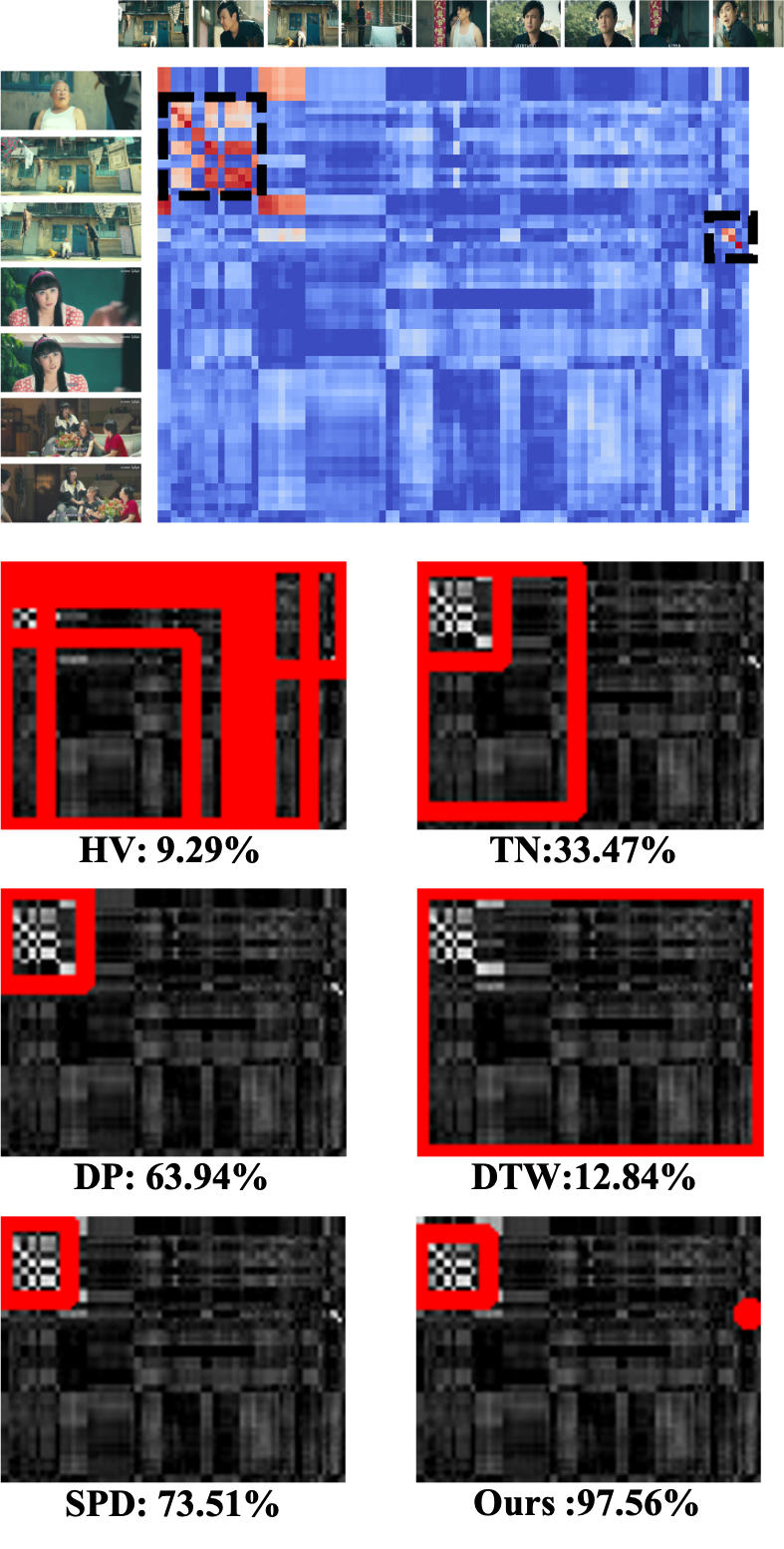}
  \caption{Example of video copy localization results. This example shows a hard case of multiple short copied segment. Only our approach can accurately search and localize both short copied segments in each video.}
  \label{sim}
\end{figure}

\begin{figure}[ht!] 
  \centering
  \includegraphics[width=0.95\linewidth]{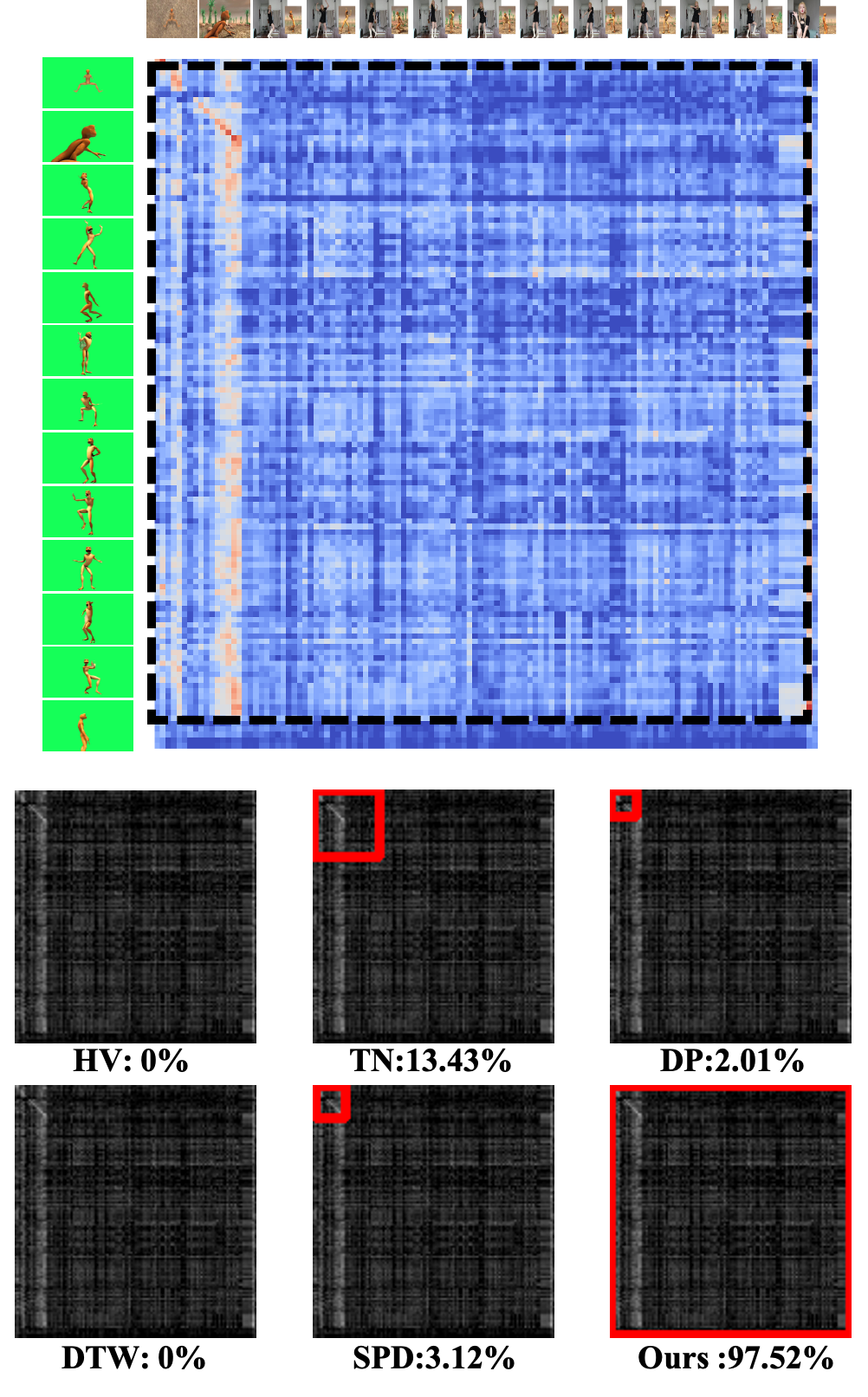}
  \caption{Example of video copy localization results. This example shows a hard case of dramatic \textit{picture in picture} frame transformation, and the background is also changed. Only our approach can accurately localize the entire copied segment in each video.}
  \label{sim}
\end{figure}

\begin{figure*}[ht!] 
  \centering
  \includegraphics[width=0.85\linewidth]{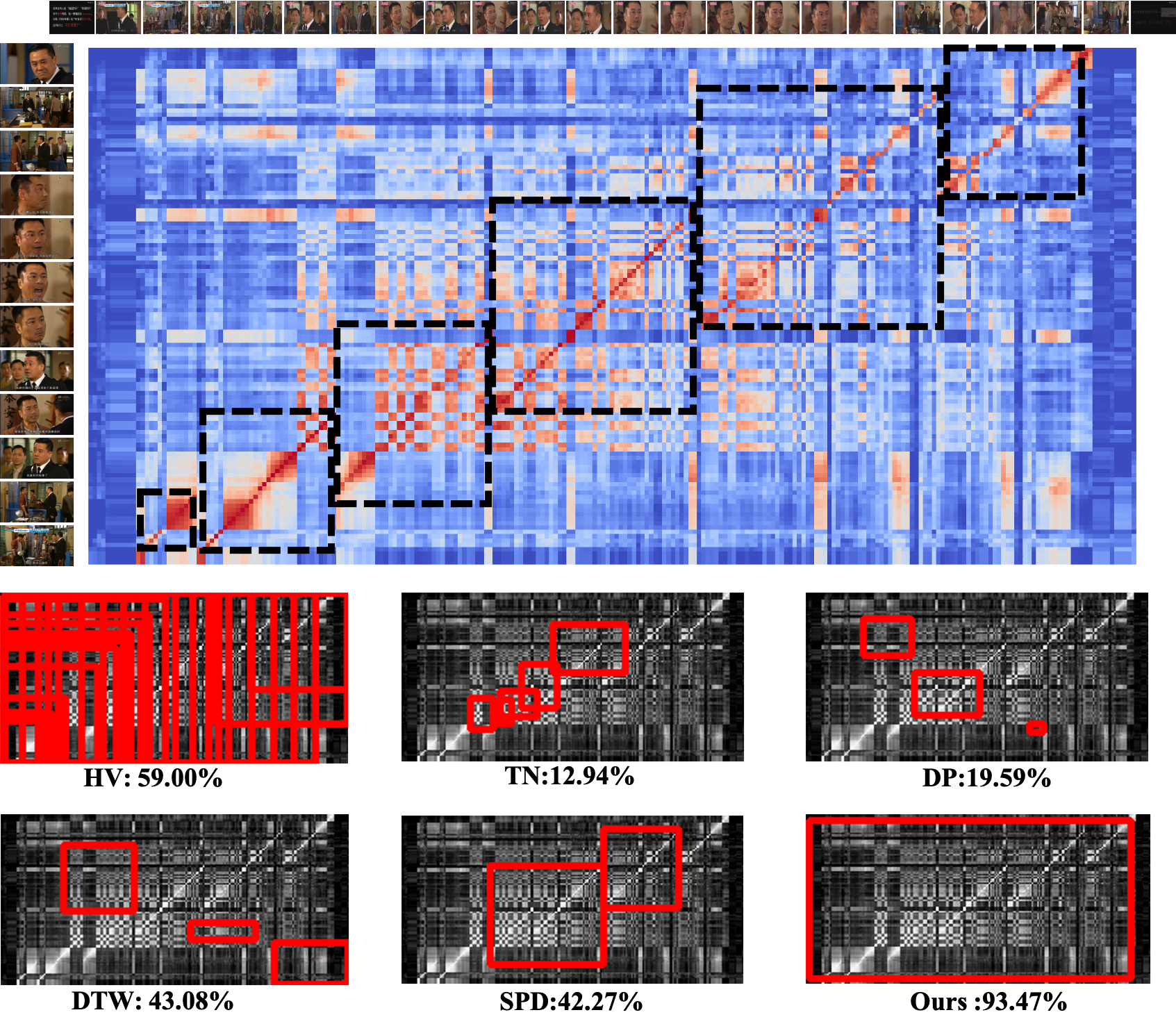}
  \caption{Example of video copy localization results. This example shows a hard case of backwards-running videos. Most of copy localization methods cannot solve this problem, and our method also shows significantly higher result.}
  \label{sim}
\end{figure*}

\begin{figure*}[ht!] 
  \centering
  \includegraphics[width=0.7\linewidth]{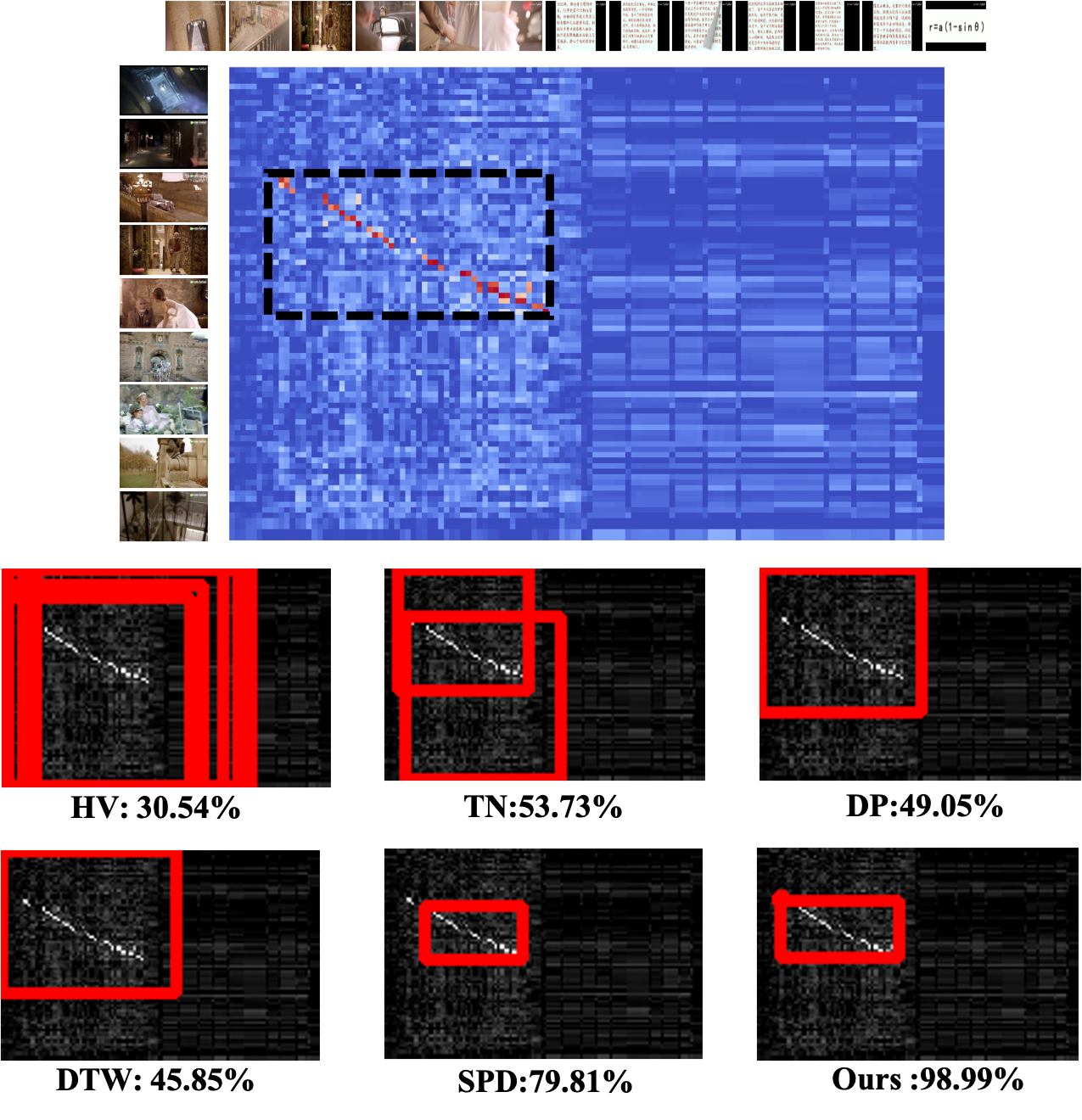}
  \caption{Example of video copy localization results. This example shows a hard case of video acceleration or deceleration. Even though most of previous methods can detect this copy transformation, they cannot output the accurate copy localization. Our approach also shows better performance and more accurate boundary.}
  \label{sim}
\end{figure*}

\bibliography{aaai23}